\documentclass[10pt,twocolumn,letterpaper]{article}

\usepackage{iccv}
\usepackage{times}
\usepackage{epsfig}
\usepackage{graphicx}
\usepackage{amsmath}
\usepackage{amssymb}
\usepackage{subfiles}
\usepackage{multirow} 
\newcommand{\ra}[1]{\renewcommand{\arraystretch}{#1}}
\usepackage{booktabs}
\usepackage[ruled]{algorithm2e}
\usepackage{bm}

\usepackage[pagebackref=true,breaklinks=true,letterpaper=true,colorlinks,bookmarks=false]{hyperref}

\iccvfinalcopy 

\def\iccvPaperID{9793} 

\ificcvfinal\fi

\begin{document}

\title{Latent Distribution Adjusting for Face Anti-Spoofing}

\author{
Qinghong Sun$^{1}$, Zhenfei Yin$^{2}$, Yichao Wu$^{2}$, Yuanhan Zhang$^{3}$, Jing Shao$^{2}$\\
$^1$Beijing University of Posts and Telecommunications, 
$^2$SenseTime Research, \\
$^3$Nanyang Technological University\\
{\tt\small qinghongsun53@gmail.com}, {\tt\small \{yinzhenfei, wuyichao, shaojing\}@sensetime.com}, {\tt\small \{yuanhan002\}@e.ntu.edu.sg} \\
}

\maketitle
\ificcvfinal\thispagestyle{empty}\fi

\begin{abstract}
With the development of deep learning, 
the field of face anti-spoofing (FAS) has witnessed great progress. FAS is usually considered a classification problem, 
where each class is assumed to contain a single cluster optimized by~\texttt{softmax} loss.
In practical deployment, 
one class can contain several local clusters,  
and a single-center is insufficient to capture the inherent structure of the FAS data.
However, 
few approaches consider large distribution discrepancies in the field of FAS.
%
In this work, 
we propose a unified framework called \textbf{Latent Distribution Adjusting (LDA)} with properties of latent, discriminative, adaptive, generic to improve the robustness of the FAS model by adjusting complex data distribution with multiple prototypes.
\textbf{1) Latent.} LDA attempts to model the data of each class as a Gaussian mixture distribution, 
and acquires a flexible number of centers for
each class in the last fully connected layer implicitly. 
\textbf{2) Discriminative.} To enhance the intra-class compactness and inter-class discrepancy, 
we propose a margin-based loss for providing distribution constrains for prototype learning. 
%
\textbf{3) Adaptive.} To make LDA more efficient and decrease redundant parameters, we propose Adaptive Prototype Selection (APS) by selecting the appropriate number of centers adaptively according to different distributions. 
%
\textbf{4) Generic.} Furthermore, 
LDA can adapt to unseen distribution by utilizing very few training data without re-training. 
Extensive experiments demonstrate that our framework can 
1) make the final representation space both intra-class compact and inter-class separable, 
2) outperform the state-of-the-art methods on multiple standard FAS benchmarks. 


\end{abstract}

\section{Introduction}

\begin{figure}[t]
\centering
\includegraphics[width=0.45\textwidth]{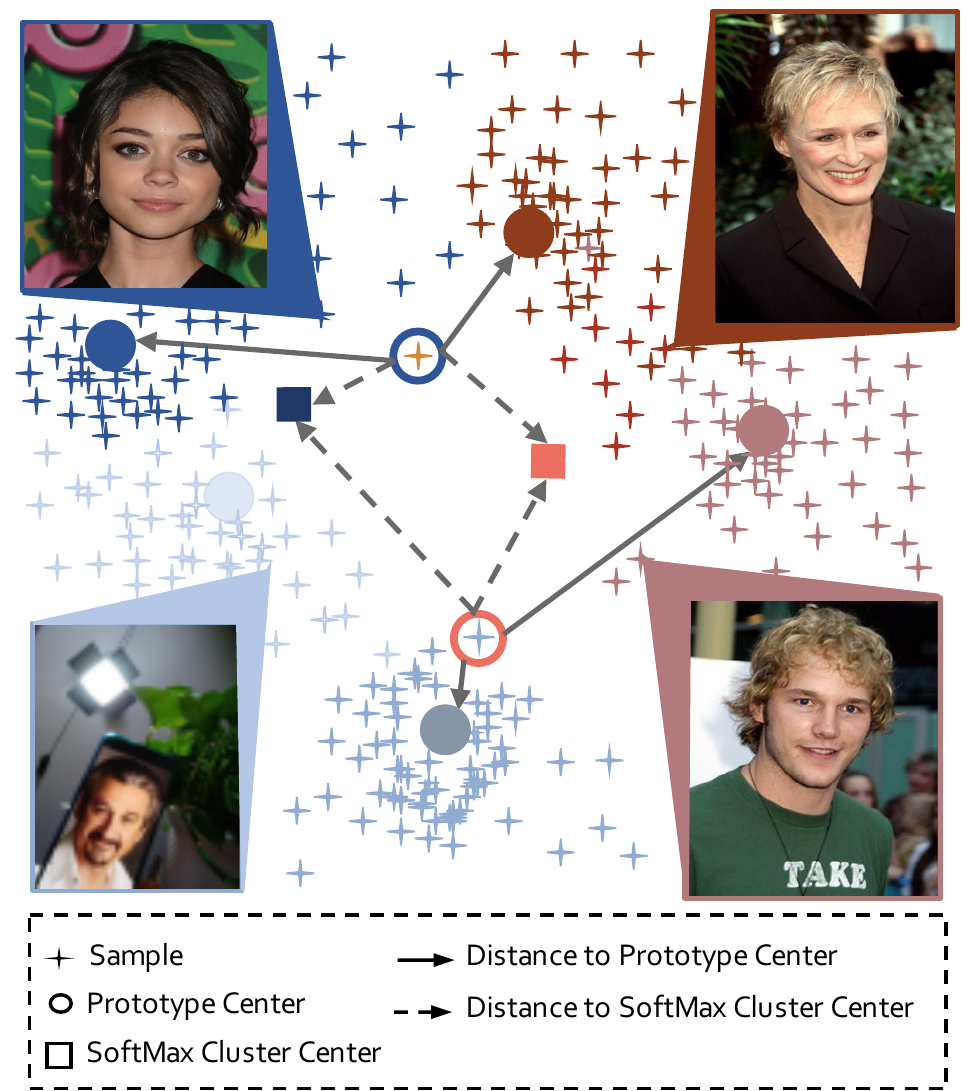}
\caption{
    The comparison between prevalent FAS method and our {Latent Distribution Adjusting} (LDA) method. %
    LDA acquires several cluster centers for each class to capture the inherent mixture distribution of the data.
    The Spoof class consists of three local clusters marked by blue. Among these clusters, the light blue cluster in the middle is comprised of several Replay Spoof samples. The other Spoof clusters capture much latent information, which is hard to be represented by simple semantic annotations. While there exist another two clusters for Live class marked by red. %
    The prevalent FAS method constrains the samples for each class only by a single center marked by squares. As a consequence, the samples enclosed by the hollow circle are predicted to the wrong class.
    However, LDA can correct these predictions by assigning several local clusters for each class. 
    }
\vspace{-6pt}
\label{figure:fig1}
\end{figure}

Face anti-spoofing (FAS), to distinguish a live face of a genuine user and a spoof face with biometric presentation attacks, is a crucial task that has a remarkable evolution~\cite{Mtt2011LBP, Komulainen2013HoG, surf, Patel2016SIFT, DoG} to ensure the security of face recognition systems. Most progress is sparked by new features and robust architectures for profiling the intrinsic property of FAS. 
Despite the many efforts, these prior works often consider FAS as a classification problem, where each class (\ie~Spoof or Live) is assumed to merely contain a single cluster optimized by the commonly used \texttt{softmax} loss. However, in the real world FAS system, each class may contain multiple interior cluster centers, whereas a single center is insufficient to capture the inherent mixture distribution of the data.
There is a surge of interest to seek for adjusting the centers of the mixture distribution to boost FAS.

A snapshot of Spoof and Live distribution\footnote{The samples are selected from CelebA-Spoof dataset~\cite{CelebA-Spoof}} is shown in Fig.~\ref{figure:fig1}, where the Spoof class marked in blue symbols has three clusters.
The light blue cluster has significant semantic representations, \ie~``Replay''.
%
While there exist two clusters for the \textit{Live} class, which are represented with red symbols. 
This toy example provides two observations: 1) A single cluster-center embedding learned by the prevailing \texttt{softmax} loss may fail in a complex data distribution as the samples enclosed by the hollow circle are wrongly predicted due to the closer distance to the wrong class. 2) Not all the clusters could be represented with semantic labels and measured by the semantic supervision. For example, it seems non-trivial to find the disciplines of semantic meaning for two respective Live clusters and two respective Spoof clusters.
%
As far as we know, few approaches consider large distribution discrepancies in the field of FAS. The most related work~\cite{SSDG}, a domain generalization method, separates the embedding feature space into $K$ Spoof clusters and one Live cluster, where $K$ is pre-defined by human prior. It is unexplainable to consider all the Live data from different domains into one cluster, and a straightforward-defined and non-learnable $K$ cannot guarantee the effectiveness of spoof classification.

Inspired by the above observations, a straightforward solution is introducing the scheme of \textit{prototype learning} (PL) by modeling complex data distribution with multiple prototypes. Prototypes represent each class with several local clusters and thus increasing the size of the last fully connected layer mildly.
The toy example in Fig.~\ref{figure:fig1} shows that with prototype learning, the measurement of the distance (in dotted line) between a sample and the \textit{class center} (in solid box) is substituted by the distance (in solid line) between the sample and the \textit{prototype center} (in the solid circle). In this way, the wrongly-predicted sample can be corrected to the right class.
Motivated by the theory of prototype learning, we propose a unified framework, named \textbf{Latent Distribution Adjusting (LDA)}, by modelling each class of FAS with multiple prototypes in a latent and adaptive way.

%
The proposed LDA is designed with a few unique properties, in comparison with the traditional PL frameworks~\cite{prototypelearning}.

\vspace{0.1cm}
\noindent (1) \textbf{Latent.} Some PL methods~\cite{HPN, DNCM} \textit{explicitly} assign different prototypes to different predefined domains or clusters in several tasks such as Few-Shot Learning~\cite{PANet}, Class-Incremental Learning~\cite{PLClass-Incremental}, \etc. 
Nevertheless, there exist predefined semantic domains labeled by human knowledge in the FAS datasets such as Spoof type, illumination condition and environment, input sensor, \etc.~\cite{CelebA-Spoof}, which causes indistinct definitions of prototypes. Therefore, LDA should assign prototypes implicitly.

\vspace{0.1cm}
\noindent (2) \textbf{Discriminative.} Traditional PL algorithms~\cite{prototypelearning, PLClass-Incremental, HPN} mainly concentrate on learning more discriminative cues. Still, for the FAS task, we focus on making the final representation for both intra-class compact and inter-class separable. In practical scenarios, the performance of FAS is measured based on thresholds (such as ACER~\cite{lyjauxuliary} and HTER~\cite{HTER}) rather than merely the classification accuracy with a threshold of 0.5 equivalently. As a consequence, we find that a more strict intra-class constraint is needed for FAS compared with general classification tasks. Accordingly, we design a margin-based loss to constrain intra-class and inter-class prototypes.

\vspace{0.1cm}
\noindent (3) \textbf{Adaptive.} For most PL methods~\cite{prototypelearning, HPN}, the numbers of prototypes are fixed during training. Due to the large distribution discrepancies of FAS data, it is difficult to manually pre-defined an appropriate prototype number in a trade-off between efficiency and effectiveness.
To this end, an \textit{Adaptive Prototype Selection} (APS) is proposed by selecting the appropriate number of prototype centers adaptively based on the data density of each prototype, and thus more data are gathered with fewer prototypes.

\vspace{0.1cm}
\noindent (4) \textbf{Generic.} With the aforementioned design, the proposed LDA has unique advantages in terms of unseen domain adaption with very few training data without retraining, which can achieve improvements with less cost.

\vspace{0.1cm}
We conduct empirical evaluations on the proposed LDA and thoroughly examine and analyze the learned prototype representations. Extensive experiments demonstrate that LDA can achieve state-of-the-art results on multiple FAS datasets and benchmarks.

\section{Related Works}

\begin{figure*}[t]
\centering
\includegraphics[width=0.95\textwidth]{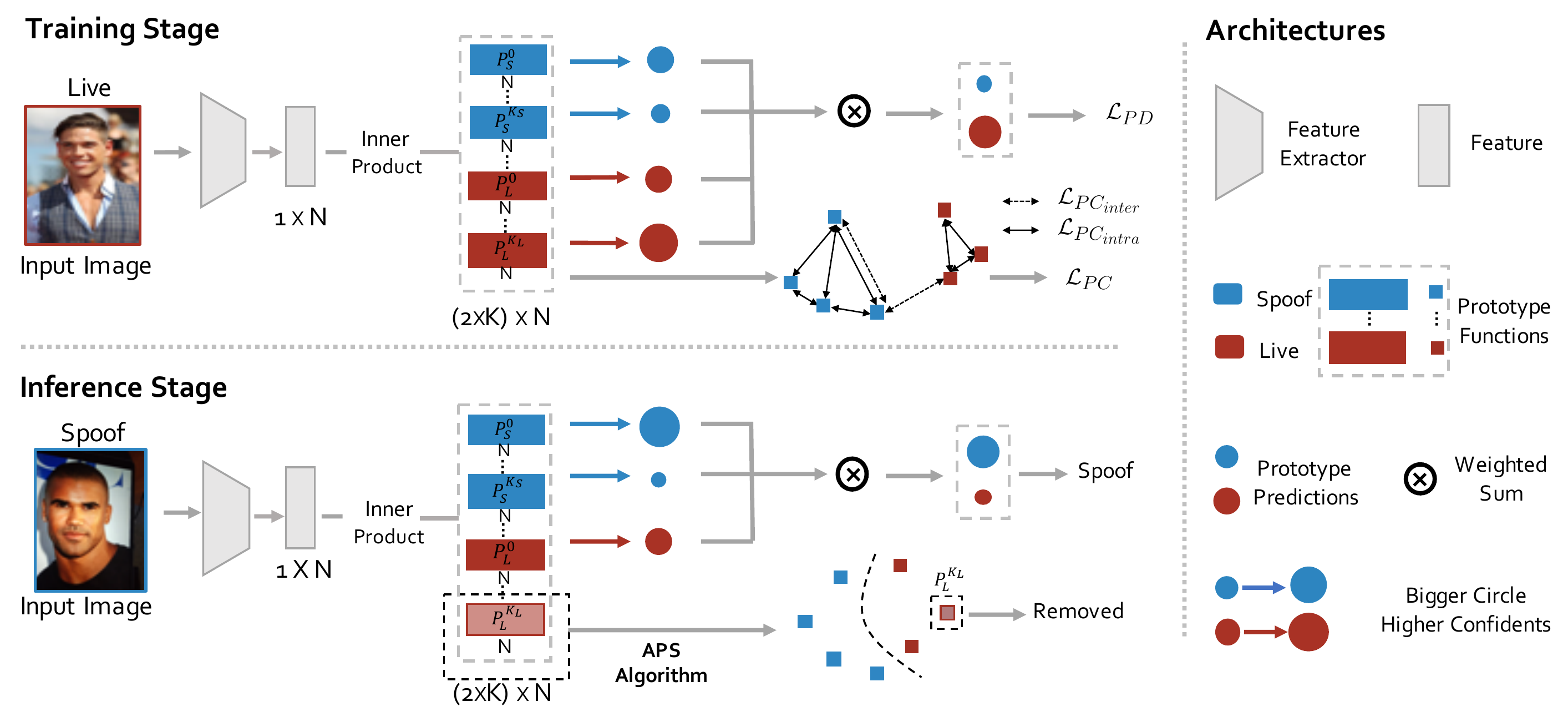}
\caption{An overview of the proposed framework LDA. LDA generates example embedding and deploys multiple learnable prototype functions in the last fully connected layer for adjusting complex data distribution. The dimension N of example embedding and prototype function is set to 512 in this paper. These embedding are fixed by $l$2 normalization. $P_{S/L}^r$ represents the $r_{th}$ prototype function from the Spoof/Live class. $K_{S/L}$ represents the number of prototype functions of the Spoof/Live class. The prototype prediction is obtained by calculating the inner product of the sample embedding and related prototype function. All of the prototype functions contribute to final decision making by a self-distributed mixture method. Prototype Center Loss ($\mathcal{L}_{PC}$) is applied for providing distribution constraints for prototype functions. The solid lines denote intra-class regularizer. Inter-class regularizer is marked by the dotted lines.
Moreover, we use Adaptive Prototype Selection (APS) algorithm in the inference stage for selecting the appropriate prototype centers adaptively.
}
\vspace{-6pt}
\label{figure:fig2}
\end{figure*}

\noindent \textbf{Face Anti-Spoofing Methods.}
Traditionally, many Face Anti-Spoofing methods adopt hand-craft features to capture the spoof cues, such as LBP~\cite{texture1,Pereira2012LBP,HTER,Mtt2011LBP}, HOG~\cite{Komulainen2013HoG,Yang2013HoG}, SURF~\cite{surf}, SIFT~\cite{Patel2016SIFT}, DoG~\cite{DoG}, etc. Some methods focused on temporal cues, such as eye-blinking~\cite{eyeblink:pan2007eyeblink, eyeblink:sun2007blinkin} and lips motion~\cite{lipmotion:kollreider2007real}. Recently, with the development of deep learning, many methods begin to employ Convolutional Neural Network(CNN) to extract discriminative cues. Yang \textit{et al.}~\cite{yang2014learn} regrades FAS as a binary classification and perform well. Atoum \textit{et al.}~\cite{AtoumFaceAU} assists the binary classification with depth map, which is learned from Fully Convolutional Network. Liu \textit{et al.}~\cite{Liu_2018_ECCV,liu20163d} leverages depth map combined with rPPG signal as the auxiliary supervision. Kim \textit{et al.}~\cite{KimBASN} utilize depth map and reflection map as the bipartite auxiliary supervision. Yang \textit{et al.}~\cite{Model_matters} combine the spatial information with global temporal information to detect attacks. And Yu \textit{et al.}~\cite{yztCDC} leverage central difference convolution to capture intrinsic detailed patterns. 
Few methods consider the distribution discrepancies. SSDG~\cite{SSDG} separate the Spoof samples while aggregate Live ones of different domains to seek a compact and generalized feature space for the Spoof class. AENet~\cite{CelebA-Spoof} constrains the distribution of embedding features of the Live and Spoof examples by multiple auxiliary centers to improve the robustness for the binary classification task.

\noindent \textbf{Prototype Learning.}
Prototype learning~\cite{prototypelearning}, claims that the lack of robustness for CNN is caused by the SoftMax layer, which is a discriminative model. It can improve the intra-class compactness of the feature representation, which can be viewed as a generative model based on the Gaussian assumption. As a flexible tool, prototype learning method~\cite{prototypelearning, HPN, DNCM} are applied to several tasks, such as Few-Shot Learning~\cite{PANet}, Zero-Shot Learning~\cite{PLZero-Shot}, Class-Incremental Learning~\cite{PLClass-Incremental}, Object Instance Search in Videos~\cite{HOPE}, etc.

\section{Methodology}


In this section, we give a detailed description of our proposed \textbf{Latent Distribution Adjusting(LDA)}. Existing works on FAS either assume each class contains a single cluster optimized by \texttt{softmax}-based loss function or manually defined clusters based on the corresponding dataset, which are insufficient to make the final representation space both intra-class compact and inter-class separable. Our method, Latent Distribution Adjusting (LDA), improves the FAS model's robustness with multiple prototypes by automatically adjusting complex data distribution with the properties of latent, discriminative, adaptive, generic.

\subsection{Overall Framework}

As shown in Fig.~\ref{figure:fig2}, the prototype functions refer to the last fully connected layer's components. LDA attempts to model data of each class as a Gaussian mixture distribution, and the prototypes act as the means of Gaussian components for each class. LDA acquires flexible numbers of efficient prototype functions for each class implicitly by forcing them to learn latent cues from image data thoroughly. All prototype functions contribute to the final class prediction. To enhance the intra-class compactness and inter-class discrepancy, we propose the 
Prototype Center Loss ($\mathcal{L}_{PC}$) to constrain the distribution of prototype centers in intra-class aspects and inter-class aspects, shown as the solid lines and dotted lines separately. After completing the training stage, we designed Adaptive Prototype Selection (APS) algorithm to adaptively and efficiently select the appropriate prototype centers for different distributions and reduce redundant parameters for LDA.

\subsection{LDA Loss}

The final objective of LDA contains two parts: a FAS classification loss based on training data and a margin-based loss function that constrains prototype centers. For the convenience of distinction, we name them as Prototype Data Loss and Prototype Center Loss respectively, \ie~ $\mathcal{L}_{PD}$ and $\mathcal{L}_{PC}$. 
%

\noindent \textbf{Prototype Data Loss.} Following conventional prototype learning~\cite{prototypelearning}, we maintain and learn multiple prototype functions in the embedding space for Live/Spoof class and use prototype matching for classification. 
We assume Spoof and Live class have equal numbers of $K$ prototypes in the initialization stage for simplicity. These prototype functions are represents as $P_j^r \in \mathbb{R}^N$, where $j \in \{0, 1\}$ represents the index of Live/Spoof class, $r \in \{1,2, ..., K\}$ represents the index of prototype functions within its class. $f_i \in \mathbf{R}^N$ denotes the embedding feature of the $i$-th sample, belonging to the $y_i$-th class.

The effectiveness of embedding normalization and weight normalization has been verified in the field of face recognition. Therefore, we utilize the normalization approach to promote the generalization ability of the LDA. 
Following~\cite{weightnorm_0, weightnorm_1, featnorm_weightnorm}, we fix prototype $\left \| P_j^r \right \| = 1$ by $l_2$ normalization. Following~\cite{featnorm_0, featnorm_weightnorm}, we fix the example embedding $\left \| f_i \right \|=1$ by $l_2$ normalization and re-scale it to $1/\tau$. $\tau$ is set to 10.0 in our experiments. After these normalization steps, the prototype predictions will be made only based on the angle between example embedding and prototype center. 

In the classification stage, samples are classified by weighted sum prototype predictions. The class $j$  prediction of the example $x_i$ is defined as follows:
\begin{equation}
\label{equation:cosine_theta_smooth}
    \cos{\theta_j} = \sum_{r=0}^K \frac{e^{\frac{1}{\tau} f_i^\top P_{j}^r}} {\sum_{r=0}^K e^{\frac{1}{\tau}f_i^\top P_{j}^r }} f_i^\top P_{j}^r.
\end{equation}

Following~\cite{weightnorm_0}, adding an angular margin penalty $m$, between $f_i$ and $P_j^r$ can increase the compactness among samples from the same class and the discrepancy among samples from different classes. By applying the class predictions to cross entropy, we define the $\mathcal{L}_{PD}$ as follows:
\begin{equation}
\label{equation:cls}
    \mathcal{L}_{PD}(x_i) = -\log \frac{e^{s(\cos({\theta_{y_i+m}}))}}
    {e^{s(\cos{(\theta_{y_i}+m))}} + e^{s(\cos{\theta_{1-y_i}})}},
\end{equation}
where $s$ is a scaling factor.

\noindent \textbf{Prototype Center Loss.} To enhance the intra-class compactness and inter-class discrepancy, we propose a margin-based Prototype Center Loss ($\mathcal{L}_{PC}$) to provide distribution constraints. $\mathcal{L}_{PC}$ consists of two components: one aims to decrease inter-class variance by guaranteeing a relative margin between the intra-class prototypes and inter-class prototypes. The other one constrains the intra-class prototype similarities by another margin penalty to reduce the intra-class variance. 
According to the observation of the prototype distribution in $\mathcal{L}_{PD}$, prototype centers from different classes may be closer than prototype centers from the same classes. The samples gathered by these prototype centers lead to the case that inter-class variation is smaller than the intra-class variation.  
Therefore, we utilize an inter-class regularizer to maintain the relationship between the inter-class variance and intra-class variance to solve this problem. The constrain is provided by adding a strict margin penalty represented as $\delta_1$ between the highest inter-class prototype similarity and the lowest intra-class prototype similarity. The loss is defined as follows:

\begin{scriptsize}
\begin{equation}
\label{equation:reg_cross}
    \mathcal{L}_{PC_{inter}}(\mathcal{P}) =  [
    \max \limits_{j, r_1, r_2}(P_{j}^{r_1} \cdot  {P_{1-j}^{r_2}}) - \min \limits_{j^{\prime},r_1^{\prime}, r_2^{\prime}}(P_{j^{\prime}}^{r_1^{\prime}} \cdot  {P_{j^{\prime}}^{r_2^{\prime}}}) + \delta_1 ] {}_{+},
\end{equation}
\end{scriptsize}
where $j, j^{\prime} \in \{0, 1\}$ represents the class. $r_1, r_2, r_1^{\prime}, r_2^{\prime}  \in \{1,2, ..., K\}$ represents the index of prototype functions for corresponding class. They subject to $r_1 \neq r_2$ and $r_1^{\prime} \neq r_2^{\prime}$. $\mathcal{P}$ represents all prototypes $\{P_j^r \}$. The plus symbol in the bottom right corner means negative values are clamped by zero. 
From our observation,
this method constrains inter-class variance between Spoof and Live class; it can develop a solution by compacting the same class prototypes. However, it decreases the effectiveness of multiple prototypes and can even degrade them to a single one. Therefore, the intra-class variance may be affected. 
To solve this problem, we propose an intra-class regularizer to reduce the whole intra-class prototype pairs' similarity. The loss is defined as follows:
\begin{equation}
\label{equation:reg_intra}
    \mathcal{L}_{PC_{intra}}(\mathcal{P}) = \sum_{j=0}^1 \sum_{r=1}^K \sum_{t=r+1}^K [P_j^{r \top} P_j^t - \delta_2]{}_{+},
\end{equation}
where $\delta_2$ is the relative margin penalty.

Integrating all modules mentioned above, the objective of the proposed LDA for FAS is:
\begin{equation}
\label{equation:loss_lda}
    \mathcal{L}_{LDA} = \mathcal{L}_{PD} + \lambda_1\mathcal{L}_{PC_{inter}} + \lambda_2\mathcal{L}_{PC_{intra}},
\end{equation}
where $\lambda_1$ and $\lambda_2$ are the balanced parameters. Therefore, LDA is end-to-end trainable.

\begin{algorithm}[t]
    \footnotesize
    \label{algo:K_adaptive}
    \SetKwFunction{FMain}{DENSITY}
            
    \SetKwProg{Fn}{Function}{:}{}
    \Fn{\FMain{$\mathcal{P}$, $\mathcal{F}$, $t$}}{
        $\mathcal{E} \gets \{\}$, $\mathcal{D} \gets \{\}$ \;
        \For{$p$ in $\mathcal{P}$}{
            $\epsilon \gets 0, \hat{\mathcal{D}} \gets \{\}$ \;
            \For{$f$ in $\mathcal{F}$}{
                \uIf{$p^\top f > t$}{
                    $\epsilon \gets \epsilon+1, \hat{\mathcal{D}} \gets \hat{\mathcal{D}} \cup f$ \;
                }
                
            }
            $\mathcal{E} \gets \mathcal{E} \cup \epsilon$ \;
            $\mathcal{D} \gets \mathcal{D} \cup \hat{\mathcal{D}}$ \;
        }
    \KwRet{$\mathcal{E}, \mathcal{D}$}\;
    }
\end{algorithm}

\subsection{Adaptive Prototype Selection(APS)}
%
%
%
In our LDA, we train LDA with equal and sufficient $K$ for Spoof and Live class. After completing the training stage for LDA, selecting appropriate prototype centers.
Inspired by the traditional DBSCAN algorithm\cite{dbscan}, the selection depends on the sample density of the relevant cluster centers. In LDA, the sample density for each prototype center is the number of samples in its region, defined by the distance threshold. As the optimization carry on in normalized embedding space, we utilize the cosine similarity to measure the distance. 
Those prototype centers with low sample density mean that they cannot gather sufficient samples in the embedding space. It means that they make few contributions to adjust relevant distribution. To remove them efficiently, we design APS algorithm to extract valid prototype with max density from the candidates continuously. 

In the initialization stage of APS, we assign one prototype center for the Live and Spoof class separately to ensure the effectiveness of binary classification. It is wasteful to cover one sample with more than one prototypes. Therefore, after popping the selected prototype from the candidates, all the samples in its region should be popped simultaneously. 
%
%
The selection process will stop when the max density of candidates is zero, or all the prototype centers are popped.
%
%
%
The detailed process of APS is shown in Algorithm~\ref{algo:K_adaptive}, where $\mathcal{F}_j, t_j$ represent the set of example embedding and density threshold for class $j$ separately. $\mathcal{P}_j^\epsilon, \mathcal{E}_j^\epsilon, \mathcal{D}_j$ represent the $\epsilon$-th prototype function, related sample density and sample set of class $j$. 
To distinguish the variables between different class, the index $0$ and $1$ are used to represent the Live and Spoof class, respectively.

\subsection{Few-shot Domain Adaptation}
\label{sec:UDA}

Traditionally, several FAS methods improve adaptability to the newly-arrived domain by utilizing domain adaption methods with unlabelled data or fine-tuning the model with labelled data,
 while LDA can effectively adapt to cross-domain data by leveraging very few labelled training data available in most practical scenarios.

We utilize one prototype for each class to demonstrate the embedding distribution of target domain data. Following ~\cite{prototypelearning}, we use the mean of each class' training data embedding from the target domain as the newly arrived prototype function. In this way, we can then directly extend the FAS method to make predictions for both the source domain and the target domain.

\subsection{Semantic Auxiliary for LDA}

Additionally, we exploit the auxiliary capacity of rich annotated semantic information for LDA. LDA$_{\mathcal{S}}$ learns with auxiliary semantic information and original prototype functions jointly. The auxiliary semantic information,  \ie~Spoof type ${\{ \mathcal{S}_k^s\}}_{k=1}^n$ and illumination conditions ${\{ \mathcal{S}_k^i\}}_{k=1}^n$ are learned via the backbone network followed by additional FC layers. 
The auxiliary supervision loss $\mathcal{L}_{Aux}$ is defined as follows:
\begin{equation}
\label{equation:loss_aux}
    \mathcal{L}_{Aux} = 
    \lambda_s\mathcal{L}_{\mathcal{S}^s} + 
    \lambda_i\mathcal{L}_{\mathcal{S}^i},
\end{equation}
where $\mathcal{L}_{\mathcal{S}^s}$ and $\mathcal{L}_{\mathcal{S}^i}$ are softmax cross entropy losses. Loss weights $\lambda_s$ and $\lambda_i$ are used to balance the contribution of each loss.
The loss function of our LDA$_{\mathcal{S}}$ is:
\begin{equation}
\label{equation:loss_ldas}
    \mathcal{L}_{LDA_{s}} = \mathcal{L}_{LDA} + 
    \lambda_{Aux}\mathcal{L}_{Aux},
\end{equation}
where $\lambda_{Aux}$ is the balanced parameter for auxiliary task. Extensive experiment results are shown in Section~\ref{sec:4.3}.


\section{Experiments}

\subsection{Experimental Settings}

\noindent \textbf{Datasets.} Three public FAS datasets are utilized with extensive experiment results to evaluate the effectiveness of our proposed methods: Oulu-NPU~\cite{oulu-npu}, SiW~\cite{lyjauxuliary} and CelebA-Spoof~\cite{CelebA-Spoof}. 

\noindent \textbf{Metrics.} As for Oulu-NPU, we follow original protocols and evaluate metrics, such as APCER, BPCER and ACER, to comparing our methods fairly. Besides, we also use TPR@FPR for evaluating in CelebA-Spoof. Moreover, Half Total Error Rate (HTER) is adopted during cross-dataset evaluation. 

\noindent \textbf{Implementation Details.}
We take ResNet-18~\cite{resnet} as the leading backbone network and pre-train it on ImageNet. The network takes face images as the input with a size of 224$\times$224. It is trained with batch size 1024 on 8 GPUs.
In Oulu-NPU experiments, the model is trained with Adam optimizer. The SGD optimizer with the momentum of 0.9 is used for CelebA-Spoof. 
Besides, detailed training procedures including learning rate and the other hyper-parameters of the loss functions are provided in the supplementary material. 

\subsection{Ablation Study}

To demonstrate the effectiveness of our LDA framework, we explore the roles of multiple prototype centers, Prototype Center Loss and Adaptive Prototype Selection (APS) algorithm. Due to the high quantity, diversity and rich annotation properties of CelebA-Spoof, relevant experiments are conducted on the intra-dataset benchmark of CelebA-Spoof with ACER metric. 

\begin{figure}[t]
\centering
\includegraphics[width=0.45\textwidth]{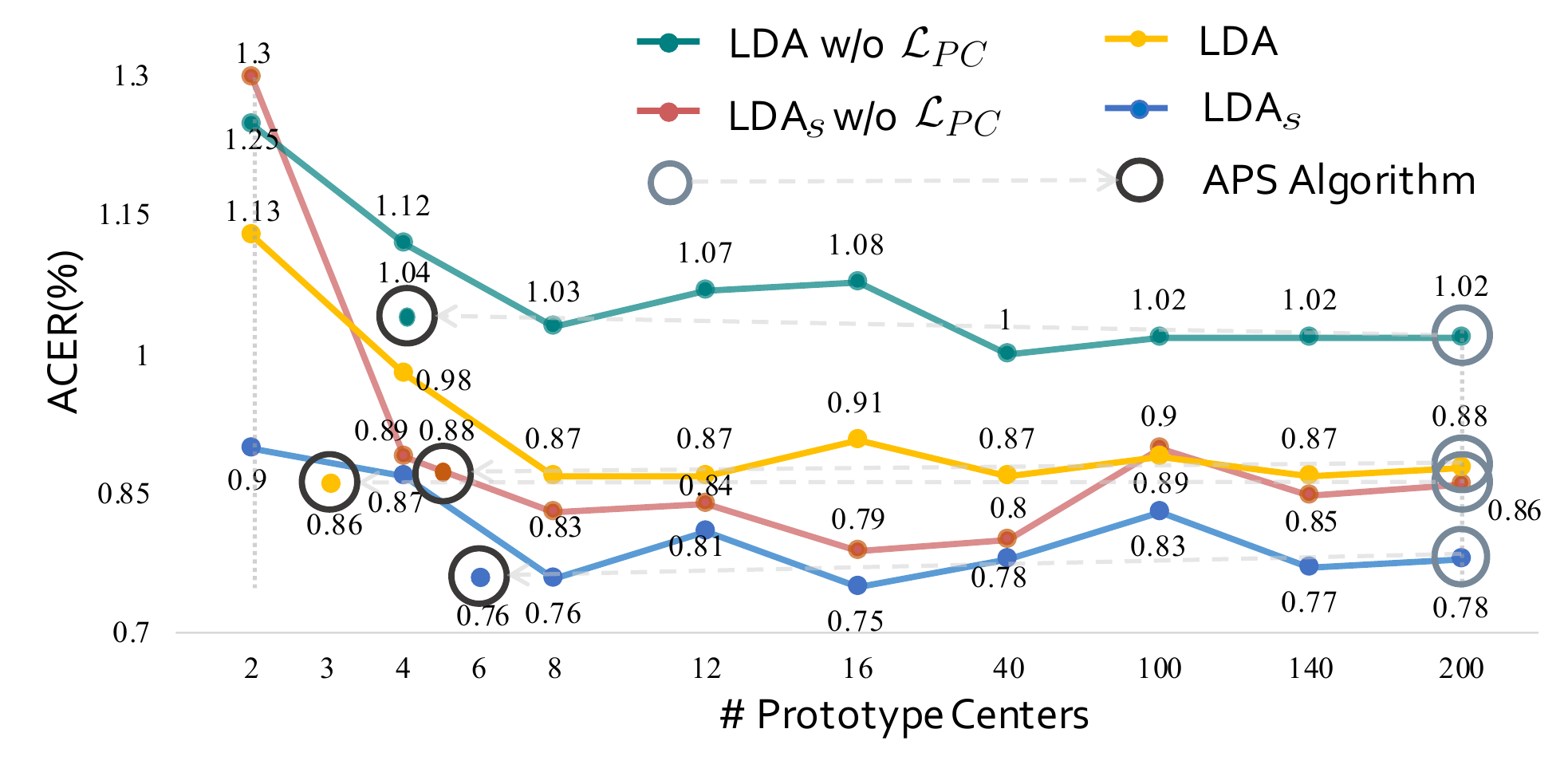}
\caption{
The number of prototype centers refers to the sum of Live ones and Spoof ones. %
The green, yellow, red and blue line refers to LDA w/o $\mathcal{L}_{PC}$,
LDA,
LDA$_{\mathcal{S}}$ w/o $\mathcal{L}_{PC}$ and LDA$_{\mathcal{S}}$ separately. %
The dots pointed by the gray circle show the performance before conducting APS algorithm. Those symbolised by the black circle show the performance and the number of selected prototype centers provided by APS algorithm. 
}
\vspace{-6pt}
\label{figure:fig3}
\end{figure}

\noindent \textbf{Implicit Prototype Learning.}
%
$\mathcal{L}_{PD}$ degenerates to general classification loss when assigning one prototype center for each class. As the green line in Fig.~\ref{figure:fig3} shows, LDA w/o $\mathcal{L}_{PC}$ has a significant improvement when increasing the number of prototype centers from 2 to 8, which confirms that $\mathcal{L}_{PD}$ is helpful to capture the hidden complex distribution for reducing the intra-class variance. 
%

\noindent \textbf{Intra-/Inter-Prototype Constrain.}
We further study about intra-class compactness and inter-class discrepancy and validate the effect of Prototype Center Loss.
Due to LDA reaches the best performance when $K$ is set to 4, our ablation experiments following this setting. Table~\ref{table:performance of different regularizers w Auxiliary Info} shows that both the inter-class module and intra-class module of the Prototype Center Loss is useful to improve the classification performance. Furthermore, the combination of these modules can further improve the performance significantly. 
Moreover, as shown in Fig.~\ref{figure:fig3}, compared with the baseline LDA w/o $\mathcal{L}_{PC}$ (green-line), both LDA with $\mathcal{L}_{PC}$ (yellow-line) and LDA with $\mathcal{L}_{Aux}$ (red-line) can achieve superior results, which demonstrate the proposed $\mathcal{L}_{PC}$ with only latent distribution adjusting (\textit{weakly supervised}) is comparable with that with auxiliary semantic annotations (\textit{fully supervised}). Furthermore, their combination (blue-line) is even better proves their complementarity.

\noindent \textbf{Number of Prototype Centers.}
As Fig.~\ref{figure:fig3} shows, the performance fluctuates in a certain range when the number of prototype centers, i.e., $K$ is over-sized. As for LDA without  $\mathcal{L}_{PC}$, four prototypes for each class is sufficient for capturing the hidden complex distribution. Nevertheless, it is hard to set a specific $K$ to deal with large discrepancies of Spoof and Live class in different datasets or applications.
APS algorithm is proposed to solve this problem. 
The circles in Fig.~\ref{figure:fig3} show the selection procedure of APS algorithm. It indicates that each class' number of prototypes can be different and adaptive. Moreover, the performance of selected prototypes is within the stable range of manual selection method. Additionally, APS can reduce redundant parameters. Accordingly, our LDA framework can adapt to various applications without manually traversing all the expected settings and over-parameterization.

\subsection{Comparison with the State-of-the-Art}
\label{sec:4.3}

\setlength{\tabcolsep}{5pt}
\begin{table}[t]
\footnotesize
\centering
\ra{1.0}
\caption{Quantitative results of $\mathcal{L}_{PC}$ ablation studies.} 
\vspace{3pt}
\label{table:performance of different regularizers w Auxiliary Info}
\begin{tabular}{@{}ccccccc@{}}
\hline
 & $\mathcal{L}_{PD}$ & $\mathcal{L}_{PC_{inter}}$  & $\mathcal{L}_{PC_{intra}}$ & ACER(\%) $\downarrow$ \\ \midrule
 & \checkmark &    &    & 1.03\\
 & \checkmark & \checkmark &   & 0.98\\  
 & \checkmark &  & \checkmark  & 0.99\\  
 & \checkmark & \checkmark &  \checkmark & 0.87 \\ 
 \bottomrule [\heavyrulewidth]
\end{tabular}%
\end{table}

\noindent \textbf{Intra-Dataset Test.}
We evaluate the intra dataset test on CelebA-Spoof. It is designed to evaluate the overall capability of the proposed method. 
As shown in Table~\ref{table:the results of intra-dataset test on CelebA-Spoof}, compared to AENet$_{\mathcal{C}, \mathcal{S}, \mathcal{G}}$, LDA improves 46.6\% for ACER and the same significant improvement for TPR@FPR, which indicates a brilliant overall capacity of LDA on a large scale dataset.
To show the effectiveness of semantic auxiliary for LDA, we conduct ablation experiments by the aid of the whole annotation information provided by  CelebA-Spoof. LDA$_{\mathcal{S}}$ outperforms LDA, which improves 13.8\% for ACER. It shows the effectiveness of rich annotated semantic for LDA.

\setlength{\tabcolsep}{5pt}
\begin{table}[t]
\centering
\ra{1.0}
\caption{Results of the intra-dataset test on CelebA-Spoof. \textbf{Bolds} are the best results. \underline{Underlines} are the second best results.}
\vspace{2pt}
\label{table:the results of intra-dataset test on CelebA-Spoof}
\resizebox{0.45\textwidth}{!}{%
\begin{tabular}{@{}ccccccc@{}}
\toprule 
\multirow{2}{*}{Methods} &
  \multicolumn{3}{c}{TPR (\%)$\uparrow$ } &

  \multirow{2}{*}{APCER (\%)$\downarrow$ } &
  \multirow{2}{*}{BPCER (\%)$\downarrow$ } &
  \multirow{2}{*}{ACER (\%)$\downarrow$ } \\ \cmidrule{2-4}
   & FPR = 1\% & FPR = 0.5\%   & FPR = 0.1\%   &        &       &           \\ \midrule
Auxiliary*~\cite{lyjauxuliary} &
97.3      & 95.2          & 83.2          & 5.71 & 1.41 & 3.56 \\ 
BASN~\cite{KimBASN}   &
  98.9 &
  \underline{97.8} &
  \textbf{90.9} &
  4.0 &
  1.1 &
  2.6 \\ 
AENet$_{\mathcal{C},\mathcal{S}, \mathcal{G}}$~\cite{CelebA-Spoof}       & 
98.9 & 
97.3          & 
87.3          & 
2.29          & 
\underline{0.96}         & 
1.63          \\
\textbf{LDA} & 
\underline{99.2} & 
97.7 & 
90.1          & 
\underline{0.58}  & 
1.17  & 
\underline{0.87}          \\ 
\textbf{LDA$_{\mathcal{S}}$} & 
\textbf{99.5} & 
\textbf{98.4} & 
\underline{90.3}          & 
\textbf{0.57}  & 
\textbf{0.94}  & 
\textbf{0.75}          \\                \bottomrule 
\end{tabular}%
}
\vspace{-4pt}
\end{table}

\noindent \textbf{Cross-Domain Test.}
The cross domain dataset test is carried out on Oulu-NPU, and CelebA-Spoof. Four protocols and two protocols are designed respectively to evaluate the generalization capability of LDA. Besides, to show the effectiveness of semantic auxiliary for LDA, we conduct ablation experiments by the aid of the annotation information from these datasets.
Oulu-NPU proposes four protocols to assess the generalization for the FAS methods. The semantic information provided by each protocol is different. In protocol I, the train and evaluation set are constructed with different sessions and the same spoof type. Therefore, we utilize the Spoof label as auxiliary information. Following this setting, the session information of protocol II, both the session information and spoof type information of protocol III are utilized for auxiliary task. 
%
As shown in Table~\ref{table:result of intra test on oulu-npu}, except for LDA$_{\mathcal{S}}$, LDA ranks the first on all four protocols of Oulu-NPU, which indicates the great generalization ability of our method on different environments conditions, spoof types, and input sensors. LDA$_{\mathcal{S}}$ outperforms LDA in two of three protocols, which improves 26.7\% and 20.0\% for ACER separately. In protocol I, compared to LDA, LDA$_{\mathcal{S}}$ causes 0.2\% decrease for ACER.
%
%
As for CelebA-Spoof, Table~\ref{table:the results of cross-domain test on CelebA-Spoof} shows that, 
compared to state-of-the-art method AENet$_{\mathcal{C}, \mathcal{S}, \mathcal{G}}$, LDA improves 46.6\% and 65.4\% on two protocols separately for ACER.
Besides, the same significant improvement is implemented for TPR@FPR. Above results indicate the great generalization capacity of LDA on larger scale dataset.
LDA$_{\mathcal{S}}$ outperforms LDA in these protocols, which improves 5.45\% and 4.67\% separately. It shows the effectiveness of rich annotated semantic for LDA.
In addition, our method achieves comparable results in SiW as shown in the supplementary material.

\setlength{\tabcolsep}{5pt}
\begin{table}[t]
\footnotesize
\centering
\ra{1.0}
\caption{Results of the cross-domain test on Oulu-NPU. \textbf{Bolds} are the best results. \underline{Underlines} are the second best results.}
\vspace{3pt}
\label{table:result of intra test on oulu-npu}
\begin{tabular}{@{}ccccc@{}}
\toprule
Prot.              & Methods        & APCER(\%)$\downarrow$ & BPCER(\%)$\downarrow$ & ACER(\%)$\downarrow$         \\ \midrule
\multirow{8}{*}{1} & GRADIANT~\cite{Gradient}     & 1.3       & 12.5      & 6.9              \\
                   & BASN~\cite{KimBASN}          & 1.5       & 5.8       & 3.6              \\
                   & Auxiliary~\cite{lyjauxuliary}     & 1.6       & 1.6       & 1.6              \\
                   & FaceDs~\cite{eccv18jourabloo}        & 1.2       & 1.7       & 1.5              \\
                   & FAS-SGTD~\cite{wang2020FAS-SGTD}         & 2.0       & 0.0       & 1.0              \\
                   & CDCN~\cite{yztCDC}           & 0.4       & 1.7       & 1.0              \\
                   & BCN~\cite{yztHMP}           & 0.0         & 1.6       & \underline{0.8}
                   \\
                    & \textbf{LDA} & 1.1 & 0.4 & \textbf{0.7}
                    \\
                    & \textbf{LDA$_{\mathcal{S}}$} & 1.6 & 0.3 & 0.9
                    \\
                   \midrule
\multirow{8}{*}{2} & FaceDs~\cite{eccv18jourabloo}        & 4.2       & 4.4       & 4.3              \\
                   & Auxiliary~\cite{lyjauxuliary}     & 2.7       & 2.7       & 2.7              \\
                   & BASN~\cite{KimBASN}          & 2.4       & 3.1       & 2.7              \\
                   & GRADIANT~\cite{Gradient}      & 3.1       & 1.9       & 2.5              \\
                   & FAS-SGTD~\cite{wang2020FAS-SGTD}         & 2.5       & 1.3       & 1.9              \\
                   & BCN~\cite{yztHMP}           & 2.6       & 0.8       & 1.7              \\
                   & CDCN~\cite{yztCDC}           & 1.5       & 1.4       & 1.5              \\
                   & \textbf{LDA} &   1.0    &   2.0     & \underline{1.5}     \\ 
                   & \textbf{LDA$_{\mathcal{S}}$} & 1.2       & 1.0       & \textbf{1.1}     \\ \midrule
\multirow{8}{*}{3} & GRADIANT~\cite{Gradient}     & 2.6$\pm $ 3.9   & 5.0$\pm $ 5.3   & 3.8$\pm $ 2.4          \\
                   & BASN~\cite{KimBASN}          & 1.8$\pm $ 1.1   & 3.5$\pm $ 3.5   & 2.7$\pm $ 1.6          \\
                   & FaceDS~\cite{eccv18jourabloo}        & 4.0$\pm $ 1.8   & 3.8$\pm $ 1.2   & 3.6$\pm $ 1.6          \\
                   & Auxuliary~\cite{lyjauxuliary}     & 2.7$\pm $ 1.3   & 3.1$\pm $ 1.7   & 2.9$\pm $ 1.5          \\
                   & FAS-SGTD~\cite{wang2020FAS-SGTD}         & 3.2$\pm $ 2.0   & 2.2$\pm $ 1.4   & 2.7$\pm $ 0.6          \\
                   & BCN~\cite{yztHMP}           & 2.8$\pm $ 2.4   & 2.3$\pm $ 2.8   & 2.5$\pm $ 1.1          \\
                   & CDCN~\cite{yztCDC}          & 2.4$\pm $ 1.3   & 2.2$\pm $ 2.0   & 2.3$\pm $ 1.4          \\
                   & \textbf{LDA} & 1.6$\pm $ 1.2   & 1.7$\pm $ 1.1   & \underline{1.5$\pm $ 1.2} \\
                   & \textbf{LDA$_{\mathcal{S}}$} & 1.3$\pm $ 0.5   & 1.0$\pm $ 1.6   & \textbf{1.2$\pm $ 1.0} \\ \midrule
\multirow{8}{*}{4} & GRADIANT~\cite{Gradient}      & 5.0$\pm $ 4.5   & 15.0$\pm $ 7.1  & 10.0$\pm $ 5.0         \\
                   & Auxiliary~\cite{lyjauxuliary}     & 9.3$\pm $ 5.6   & 10.4$\pm $ 6.0  & 9.5$\pm $ 6.0          \\       
                    & CDCN~\cite{yztCDC}           & 4.6$\pm $ 4.6   & 9.2$\pm $ 8.0   & 6.9$\pm $ 2.9          \\
                   & FaceDS~\cite{eccv18jourabloo}        & 1.2$\pm $ 6.3   & 6.1$\pm $ 5.11  & 5.6$\pm $ 5.7          \\
                   & BASN~\cite{KimBASN}          & 6.4$\pm $ 8.6   & 7.5$\pm $ 6.9   & 5.2$\pm $ 3.7          \\
                   & BCN~\cite{yztHMP}           & 2.9$\pm $ 4.0   & 7.5$\pm $ 6.9   & 5.2$\pm $ 3.7          \\
                   & FAS-SGTD~\cite{wang2020FAS-SGTD}         & 6.7$\pm $ 7.5   & 3.3$\pm $ 4.1   & \underline{5.0$\pm $ 2.2}          \\
                   & \textbf{LDA} & 2.1$\pm $ 2.2  & 3.9$\pm $ 5.7   & \textbf{2.7$\pm $ 3.3} \\  \bottomrule
\end{tabular}%
\vspace{-9pt}
\end{table}

\setlength{\tabcolsep}{5pt}
\begin{table}[t]
\centering
\ra{1.0}
\caption{Results of the cross-domain test on CelebA-Spoof. \textbf{Bolds} are the best results. \underline{Underlines} are the second best results.}
\vspace{3pt}
\label{table:the results of cross-domain test on CelebA-Spoof}
\resizebox{0.45\textwidth}{!}{%
\begin{tabular}{cccccccccc}
\toprule
\multirow{2}{*}{Prot.} &
  \multirow{2}{*}{Methods} &
  \multicolumn{3}{c}{TPR (\%) $\uparrow$} &
  \multirow{2}{*}{APCER (\%)$\downarrow$} &
  \multirow{2}{*}{BPCER (\%)$\downarrow$} &
  \multirow{2}{*}{ACER (\%)$\downarrow$} \\ \cmidrule{3-5}
 &
   &
  FPR = 1\% &
  FPR = 0.5\% &
  FPR = 0.1\% &
   &
   &
   \\ \midrule
   
\multirow{3}{*}{1} 
&
AENet$_{\mathcal{C}, \mathcal{S}, \mathcal{G}}$~\cite{CelebA-Spoof} &
  95.0 &
  91.4 &
  73.6 &
  4.09 &
  2.09 &
  3.09  \\ 
& \textbf{LDA}&
  \underline{96.9} &
  \textbf{94.3} &
  \textbf{81.7} &
  \textbf{1.41} &
  \underline{1.89} &
  \underline{1.65}  \\ 
& \textbf{LDA$_{\mathcal{S}}$}&
  \textbf{97.3} &
  \underline{94.1} &
  \underline{81.3} &
  \underline{1.82} &
  \textbf{1.30} &
  \textbf{1.56}  \\  \midrule

\multirow{3}{*}{2} 
&
AENet$_{\mathcal{C}, \mathcal{S}, \mathcal{G}}$~\cite{CelebA-Spoof} &
  
  \# &
  \# &
  \# &
  4.94$\pm $3.42 &
  1.24$\pm $0.73 &
  3.09$\pm $2.08  \\
& \textbf{LDA} &
  \# &
  \# &
  \# &
  \textbf{0.98$\pm $0.35} &
  \underline{1.14$\pm $0.38} &
  \underline{1.07$\pm $0.36}  \\
& \textbf{LDA$_{\mathcal{S}}$} &
  \# &
  \# &
  \# &
  \underline{1.09$\pm $0.46} &
  \textbf{0.95$\pm $0.30} &
  \textbf{1.02$\pm $0.38}  \\
  \bottomrule
\end{tabular}%
}
\vspace{-5pt}
\end{table}


\begin{figure*}[t]
\centering
\includegraphics[width=0.98\textwidth]{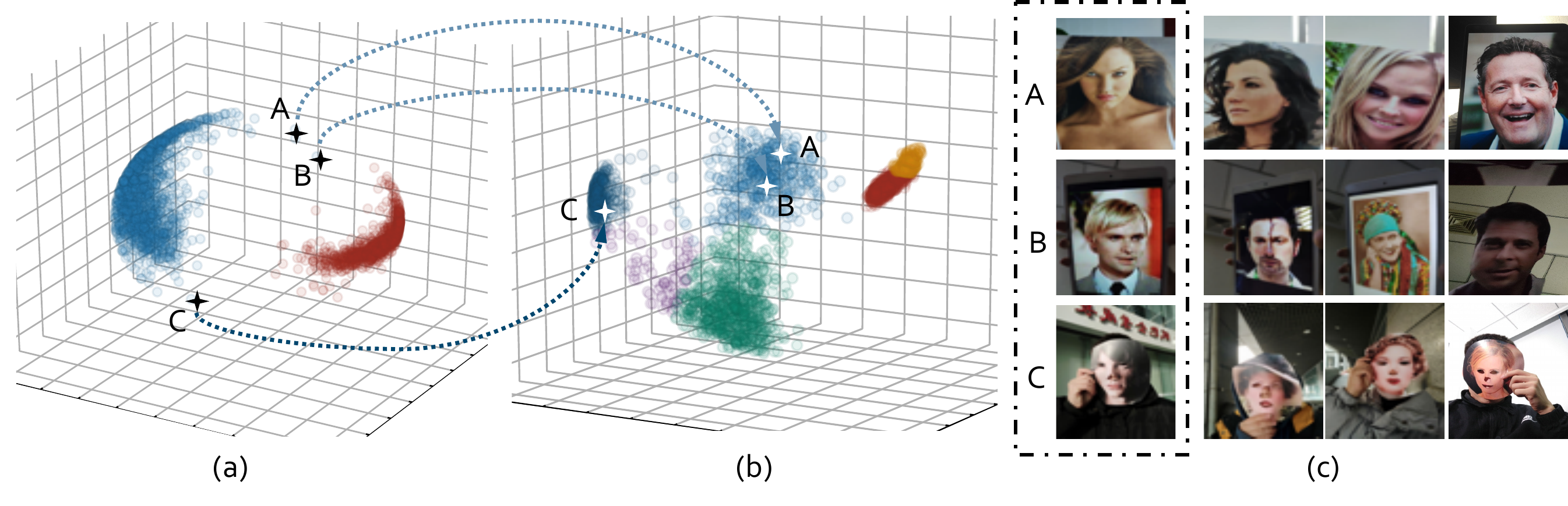}
\caption{ Comparison between prevalent FAS method and our LDA method in real embedding space. (a) and (b) are the sample distributions of prevalent FAS method and LDA separately. 
(c) visualizes some samples from these distributions.
The first column symbolised by  dot rectangle shows the outliers of prevalent FAS method in its embedding space~(a). The dot lines show their distribution discrepancy within different embedding spaces. For each sample, the other samples within the same row are its neighbours in LDA embedding space.
}
\vspace{-6pt}
\label{figure:fig5}
\end{figure*}

\setlength{\tabcolsep}{5pt}
\begin{table}[htb]
\centering
\ra{1.0}
\caption{Results of cross-dataset testing between CelebA-Spoof and SiW. \textbf{Bolds} are the best results.}
\vspace{3pt}
\label{table:Adaptation to new domain with few training data}
\resizebox{0.45\textwidth}{!}{%
\begin{tabular}{@{}ccccccc@{}}
\toprule
 & Methods & Train  & APCER(\%) $\downarrow$ & BPCER(\%) $\downarrow$ & ACER(\%) $\downarrow$ & HTER(\%) $\downarrow$ \\ \midrule
 \multirow{2}{*}{(1)} & ResNet-18 & A   &  1.04  & 1.31 & 1.17 & 21.89 \\
  & \textbf{LDA} & A   &  \textbf{0.69}  & \textbf{0.84} & \textbf{0.76} & \textbf{17.80} \\ \hline
 \multirow{2}{*}{(2)} & ResNet-18 & A \& $B_1$ &  $1.13\pm 0.04$  & $1.34\pm 0.09$ & $1.22\pm 0.09$ & $20.71\pm 1.55$ \\
 & \textbf{LDA} & $B_1$ & \bm{$0.48\pm 0.03$}  & \bm{$0.92\pm 0.03$} & \bm{$0.70\pm 0.02$} & \bm{$16.93\pm 0.45$}\\ \hline
 \multirow{2}{*}{(3)} & ResNet-18 & A \& $B_2$ &  $0.95\pm 0.10$  & $1.53\pm 0.08$ & $1.24\pm 0.06$ & $20.50\pm 1.90$ \\
 & \textbf{LDA} & $B_2$ &  \bm{$0.49\pm 0.03$}  & \bm{$0.85\pm 0.02$} & \bm{$0.68\pm 0.01$} & \bm{$16.13\pm 0.16$} \\
 \bottomrule
\end{tabular}%
}
\vspace{-12pt}
\end{table}

\noindent \textbf{Cross-Dataset Test.}
To further evaluate the generalization ability of LDA based on practical scenarios, we conduct cross dataset test with two protocols. One is training on the CelebA-Spoof and testing on the SiW. As Table~\ref{table:Adaptation to new domain with few training data} (1) shows, LDA outperforms traditional binary supervision method (ResNet-18 with \texttt{softmax}) in terms of ACER and HTER, which demonstrates its strong direct adaptability for unseen domain data.
%
%
The second protocol is a few labelled training data from the unseen domain is available for adaption. This protocol is used to evaluate the adaptability of LDA by following practical scenarios, which is mentioned as~\ref{sec:UDA}. We regard CelebA-Spoof intra-dataset as the known domain called A and SiW intra-dataset as the target domain. We sample very few training samples from SiW intra-dataset as the training data from the target domain. To demonstrate the adaptive performance of our method fairly, we experiment with different sample sizes, such as 30 and 300. Related training data is represented by $B_1$ and $B_2$, respectively. These training data is randomly sampled. To ensure the credibility of experimental results, we conduct five experiments for each sample size and take the average performance of them as the final performance. 
In this setting, traditional binary supervision (ResNet-18) gather previous training data and a few training data from the target domain as the new training set to fine-tune the model.
To further improve the performance of fine-tune method, we try to upsample the few training data from the target domain and increase relevant loss weight for leveraging the few data sufficiently. However, increasing the effect of few target domain data cannot promote the adaptability in this case.
As for LDA, we follow the adaptation method mentioned as~\ref{sec:UDA}.
The performance comparison between binary supervision fine-tune method and LDA are shown as Table~\ref{table:Adaptation to new domain with few training data}. As (1) and (2) show, compared with evaluating on target domain directly, both the fine-tune method and LDA are useful for adapting to the target domain. 
As (2) shows, compared to the fine-tune method, LDA has better performance both in the known domain and target domain. (3) shows that, as the sample size increases, both the fine-tune method and LDA achieve better performance. 
%
Furthermore, the effectiveness of LDA is more significant when obtaining more target domain training data. 
%

\subsection{Further Analysis}
\begin{figure}[t]
\centering
\includegraphics[width=0.45\textwidth]{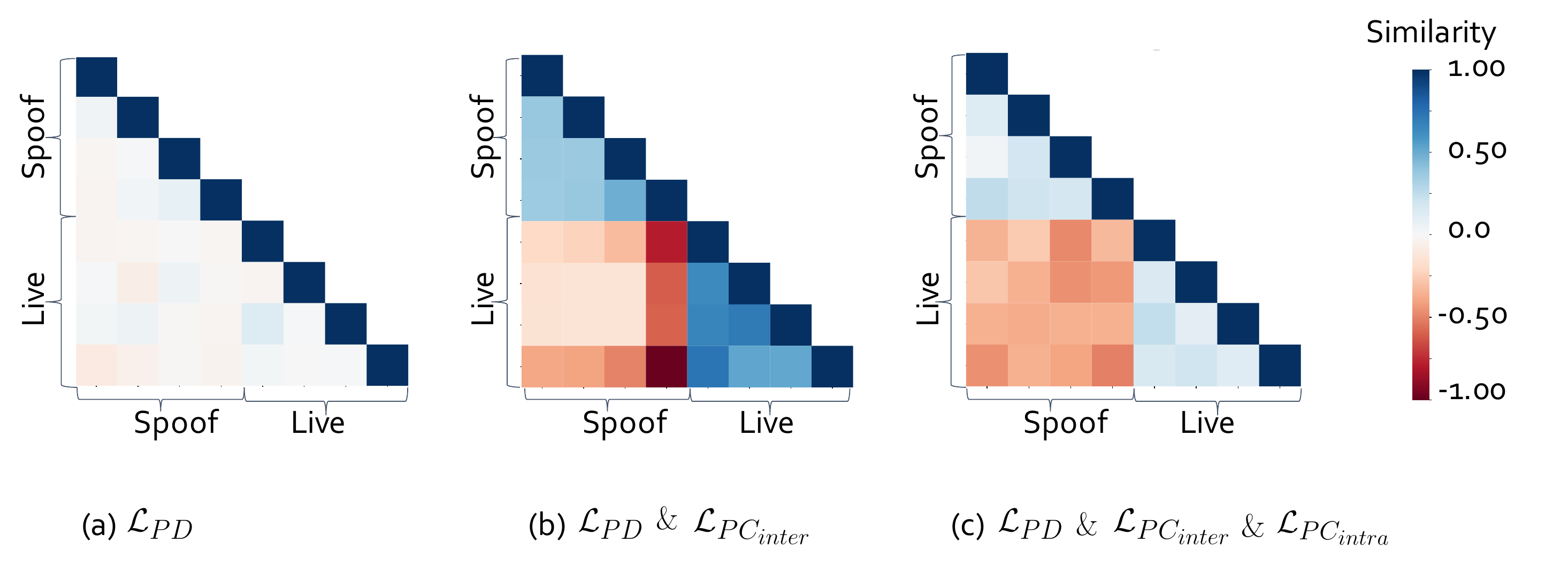}
\caption{The comparison of the prototype similarities driven by the modules of Prototype Center Loss. The Live/Spoof annotation represents the class of corresponding prototypes. Because the diagonal elements correspond to the same prototype on the horizontal and vertical axes, all the values are equal to one.
}
\vspace{-10pt}
\label{figure:fig4}
\end{figure}

\noindent \textbf{Visualization of the Prototype Similarities.} In order to observe and explore the effectiveness of Prototype Center Loss, we conduct ablation study and visualize the similarities among prototypes. As Fig.~\ref{figure:fig4} (a) shows, all of the similarities are very close. Therefore, the significant diversity between the intra-class variance and inter-class variance does not exist. Inter-class module can separate inter-class prototype effectively as shown in Fig.~\ref{figure:fig4} (b). However, this improvement is achieved by sacrificing the discrepancy for intra-class prototypes. It leads to a decrease of adjustability for LDA. This problem can be solved by introducing intra-class module, which can also maintain inter-class prototype distribution simultaneously as shown in Fig.~\ref{figure:fig4} (c).
%

\noindent \textbf{Visualization of the Embedding Space.} To demonstrate the embedding space learned by LDA, we adopt t-SNE~\cite{tsne} to show the comparison between prevalent FAS method and our Latent Distribution Adjusting (LDA) method. Fig.~\ref{figure:fig5} (a) shows the performance of prevalent FAS method. Each class is represented by a single cluster. Spoof samples is more dispersed than Live samples. There exist some outliers, which cannot be constrained well by these clusters. Therefore, outliers can be classified into the wrong class easily. LDA can solve this problem by introducing multiple local clusters. As Fig.~\ref{figure:fig5} (b) shows that there are six clusters, two for Live class and four for Spoof class. Compared to the prevalent FAS method, the Live samples are more compact, and the Spoof class is represented by several local clusters with low intra-class variance.
We sample some outliers of the prevalent FAS method shown as the first column of Fig.~\ref{figure:fig5} (c). For each example, the other examples within the same row are its neighbours in LDA. As the dot lines in Fig.~\ref{figure:fig5} (a) and (b) show, the outliers in the prevalent FAS method are contrained well by local clusters in LDA. Besides, the neighbor examples of these examples have significant semantic commons. Above all, LDA does well in these outliers and is able to learn some semantic pattern. 


\section{Conclusion}
In this work, we observe and analyze the large distribution discrepancies in the field of FAS. We propose a unified framework called Latent Distribution Adjusting (LDA) to improve the robustness of the FAS model by adjusting complex data distribution with multiple prototype centers.
To enhance the intra-class compactness and inter-class discrepancy, 
we propose a margin-based loss for providing distribution constrains for prototype learning. 
To make LDA more efficient and decrease redundant parameters, 
we propose Adaptive Prototype Selection (APS) to select the appropriate prototype centers adaptively according to different domains. 
Furthermore, 
LDA can adapt to unseen distribution effectively by utilizing very few training data without re-training. 
Extensive experimental results on multiple standard FAS benchmarks demonstrate the robustness of proposed LDA framework.

\clearpage
{\small
\bibliographystyle{ieee_fullname}
\bibliography{ICCV_LDA/sections/egbib}

\begin{thebibliography}{10}

\bibitem{AtoumFaceAU}
Yousef Atoum, Yaojie Liu, Amin Jourabloo, and Xiaoming Liu.
\newblock Face anti-spoofing using patch and depth-based cnns.
\newblock In {\em IJCB}, pages 319--328. IEEE, 2017.

\bibitem{Gradient}
Zinelabdine Boulkenafet, Jukka Komulainen, Zahid Akhtar, Azeddine Benlamoudi,
  Djamel Samai, Salah~Eddine Bekhouche, Abdelkrim Ouafi, Fadi Dornaika,
  Abdelmalik Taleb-Ahmed, Le~Qin, et~al.
\newblock A competition on generalized software-based face presentation attack
  detection in mobile scenarios.
\newblock In {\em IJCB}, pages 688--696. IEEE, 2017.

\bibitem{texture1}
Zinelabidine Boulkenafet, Jukka Komulainen, and Abdenour Hadid.
\newblock Face anti-spoofing based on color texture analysis.
\newblock In {\em ICIP}, pages 2636--2640. IEEE, 2015.

\bibitem{surf}
Zinelabidine Boulkenafet, Jukka Komulainen, and Abdenour Hadid.
\newblock Face antispoofing using speeded-up robust features and fisher vector
  encoding.
\newblock {\em IEEE Signal Processing Letters}, 24(2):141--145, 2016.

\bibitem{HSV:boulkenafet2016face}
Zinelabidine Boulkenafet, Jukka Komulainen, and Abdenour Hadid.
\newblock Face spoofing detection using colour texture analysis.
\newblock {\em TIFS}, 11(8):1818--1830, 2016.

\bibitem{oulu-npu}
Zinelabinde Boulkenafet, Jukka Komulainen, Lei Li, Xiaoyi Feng, and Abdenour
  Hadid.
\newblock Oulu-npu: A mobile face presentation attack database with real-world
  variations.
\newblock In {\em IEEE International Conference on Automatic Face \& Gesture
  Recognition}, pages 612--618. IEEE, 2017.

\bibitem{replay_attack}
Ivana Chingovska, Andr{\'e} Anjos, and S{\'e}bastien Marcel.
\newblock On the effectiveness of local binary patterns in face anti-spoofing.
\newblock In {\em BIOSIG}, pages 1--7. IEEE, 2012.

\bibitem{xception}
Fran{\c{c}}ois Chollet.
\newblock Xception: Deep learning with depthwise separable convolutions.
\newblock In {\em CVPR}, pages 1251--1258, 2017.

\bibitem{Pereira2012LBP}
Tiago de~Freitas~Pereira, Andr{\'e} Anjos, Jos{\'e}~Mario De~Martino, and
  S{\'e}bastien Marcel.
\newblock Lbp- top based countermeasure against face spoofing attacks.
\newblock In {\em ACCV}, pages 121--132. Springer, 2012.

\bibitem{HTER}
Tiago de~Freitas~Pereira, Andr{\'e} Anjos, Jos{\'e}~Mario De~Martino, and
  S{\'e}bastien Marcel.
\newblock Can face anti-spoofing countermeasures work in a real world scenario?
\newblock In {\em ICB}, pages 1--8. IEEE, 2013.

\bibitem{weightnorm_0}
Jiankang Deng, Jia Guo, and Stefanos~Zafeiriou Arcface.
\newblock Additive angular margin loss for deep face recognition.
\newblock {\em CoRR}, 2018.

\bibitem{dbscan}
Martin Ester, Hans-Peter Kriegel, J{\"o}rg Sander, and Xiaowei Xu.
\newblock Density-based spatial clustering of applications with noise.
\newblock In {\em Int. Conf. Knowledge Discovery and Data Mining}, volume 240,
  page~6, 1996.

\bibitem{DNCM}
Samantha Guerriero, Barbara Caputo, and Thomas Mensink.
\newblock Deep nearest class mean classifiers.
\newblock In {\em International Conference on Learning Representations,
  Worskhop Track}, 2018.

\bibitem{resnet}
Kaiming He, Xiangyu Zhang, Shaoqing Ren, and Jian Sun.
\newblock Deep residual learning for image recognition.
\newblock In {\em CVPR}, pages 770--778, 2016.

\bibitem{SSDG}
Yunpei Jia, Jie Zhang, Shiguang Shan, and Xilin Chen.
\newblock Single-side domain generalization for face anti-spoofing.
\newblock In {\em CVPR}, June 2020.

\bibitem{eccv18jourabloo}
Amin Jourabloo, Yaojie Liu, and Xiaoming Liu.
\newblock Face de-spoofing: Anti-spoofing via noise modeling.
\newblock In {\em ECCV}, pages 290--306, 2018.

\bibitem{KimBASN}
Taewook Kim, YongHyun Kim, Inhan Kim, and Daijin Kim.
\newblock Basn: Enriching feature representation using bipartite auxiliary
  supervisions for face anti-spoofing.
\newblock In {\em ICCVW}, pages 0--0, 2019.

\bibitem{lipmotion:kollreider2007real}
Klaus Kollreider, Hartwig Fronthaler, Maycel~Isaac Faraj, and Josef Bigun.
\newblock Real-time face detection and motion analysis with application in
  “liveness” assessment.
\newblock {\em TIFS}, 2(3):548--558, 2007.

\bibitem{Komulainen2013HoG}
Jukka Komulainen, Abdenour Hadid, and Matti Pietik{\"a}inen.
\newblock Context based face anti-spoofing.
\newblock In {\em BTAS}, pages 1--8. IEEE, 2013.

\bibitem{Fourier:li2004live}
Jiangwei Li, Yunhong Wang, Tieniu Tan, and Anil~K Jain.
\newblock Live face detection based on the analysis of fourier spectra.
\newblock In {\em Biometric technology for human identification}, volume 5404,
  pages 296--303. International Society for Optics and Photonics, 2004.

\bibitem{Liu_2018_ECCV}
Si-Qi Liu, Xiangyuan Lan, and Pong~C Yuen.
\newblock Remote photoplethysmography correspondence feature for 3d mask face
  presentation attack detection.
\newblock In {\em ECCV}, pages 558--573, 2018.

\bibitem{liu20163d}
Siqi Liu, Pong~C Yuen, Shengping Zhang, and Guoying Zhao.
\newblock 3d mask face anti-spoofing with remote photoplethysmography.
\newblock In {\em ECCV}, pages 85--100. Springer, 2016.

\bibitem{weightnorm_1}
Weiyang Liu, Yandong Wen, Zhiding Yu, Ming Li, Bhiksha Raj, and Le~Song.
\newblock Sphereface: Deep hypersphere embedding for face recognition.
\newblock In {\em CVPR}, pages 212--220, 2017.

\bibitem{lyjauxuliary}
Yaojie Liu, Amin Jourabloo, and Xiaoming Liu.
\newblock Learning deep models for face anti-spoofing: Binary or auxiliary
  supervision.
\newblock In {\em ICCV}, pages 389--398, 2018.

\bibitem{tsne}
Laurens van~der Maaten and Geoffrey Hinton.
\newblock Visualizing data using t-sne.
\newblock {\em Journal of machine learning research}, 9(Nov):2579--2605, 2008.

\bibitem{Mtt2011LBP}
Jukka M{\"a}{\"a}tt{\"a}, Abdenour Hadid, and Matti Pietik{\"a}inen.
\newblock Face spoofing detection from single images using micro-texture
  analysis.
\newblock In {\em IJCB}, pages 1--7. IEEE, 2011.

\bibitem{HPN}
Pascal Mettes, Elise van~der Pol, and Cees Snoek.
\newblock Hyperspherical prototype networks.
\newblock In {\em NIPS}, pages 1487--1497, 2019.

\bibitem{eyeblink:pan2007eyeblink}
Gang Pan, Lin Sun, Zhaohui Wu, and Shihong Lao.
\newblock Eyeblink-based anti-spoofing in face recognition from a generic
  webcamera.
\newblock In {\em ICCV}, pages 1--8. IEEE, 2007.

\bibitem{Patel2016SIFT}
Keyurkumar Patel, Hu~Han, and Anil~K Jain.
\newblock Secure face unlock: Spoof detection on smartphones.
\newblock {\em TIFS}, 11(10):2268--2283, 2016.

\bibitem{DoG}
Bruno Peixoto, Carolina Michelassi, and Anderson Rocha.
\newblock Face liveness detection under bad illumination conditions.
\newblock In {\em ICIP}, pages 3557--3560. IEEE, 2011.

\bibitem{featnorm_0}
Rajeev Ranjan, Carlos~D Castillo, and Rama Chellappa.
\newblock L2-constrained softmax loss for discriminative face verification.
\newblock {\em arXiv preprint arXiv:1703.09507}, 2017.

\bibitem{eyeblink:sun2007blinkin}
Lin Sun, Gang Pan, Zhaohui Wu, and Shihong Lao.
\newblock Blinking-based live face detection using conditional random fields.
\newblock In {\em ICB}, pages 252--260. Springer, 2007.

\bibitem{featnorm_weightnorm}
Feng Wang, Jian Cheng, Weiyang Liu, and Haijun Liu.
\newblock Additive margin softmax for face verification.
\newblock {\em IEEE Signal Processing Letters}, 25(7):926--930, 2018.

\bibitem{PANet}
Kaixin Wang, Jun~Hao Liew, Yingtian Zou, Daquan Zhou, and Jiashi Feng.
\newblock Panet: Few-shot image semantic segmentation with prototype alignment.
\newblock In {\em ICCV}, October 2019.

\bibitem{wang2020FAS-SGTD}
Zezheng Wang, Zitong Yu, Chenxu Zhao, Xiangyu Zhu, Yunxiao Qin, Qiusheng Zhou,
  Feng Zhou, and Zhen Lei.
\newblock Deep spatial gradient and temporal depth learning for face
  anti-spoofing.
\newblock In {\em CVPR}, pages 5042--5051, 2020.

\bibitem{prototypelearning}
Hong-Ming Yang, Xu-Yao Zhang, Fei Yin, and Cheng-Lin Liu.
\newblock Robust classification with convolutional prototype learning.
\newblock In {\em CVPR}, June 2018.

\bibitem{yang2014learn}
Jianwei Yang, Zhen Lei, and Stan~Z Li.
\newblock Learn convolutional neural network for face anti-spoofing.
\newblock {\em arXiv preprint arXiv:1408.5601}, 2014.

\bibitem{Yang2013HoG}
Jianwei Yang, Zhen Lei, Shengcai Liao, and Stan~Z Li.
\newblock Face liveness detection with component dependent descriptor.
\newblock In {\em ICB}, pages 1--6. IEEE, 2013.

\bibitem{Model_matters}
Xiao Yang, Wenhan Luo, Linchao Bao, Yuan Gao, Dihong Gong, Shibao Zheng,
  Zhifeng Li, and Wei Liu.
\newblock Face anti-spoofing: Model matters, so does data.
\newblock In {\em CVPR}, pages 3507--3516, 2019.

\bibitem{PLClass-Incremental}
Lu~Yu, Bartlomiej Twardowski, Xialei Liu, Luis Herranz, Kai Wang, Yongmei
  Cheng, Shangling Jui, and Joost van~de Weijer.
\newblock Semantic drift compensation for class-incremental learning.
\newblock In {\em CVPR}, June 2020.

\bibitem{HOPE}
Tan Yu, Yuwei Wu, and Junsong Yuan.
\newblock Hope: Hierarchical object prototype encoding for efficient object
  instance search in videos.
\newblock In {\em CVPR}, July 2017.

\bibitem{yztHMP}
Zitong Yu, Xiaobai Li, Xuesong Niu, Jingang Shi, and Guoying Zhao.
\newblock Face anti-spoofing with human material perception.
\newblock {\em arXiv preprint arXiv:2007.02157}, 2020.

\bibitem{yztCDC}
Zitong Yu, Chenxu Zhao, Zezheng Wang, Yunxiao Qin, Zhuo Su, Xiaobai Li, Feng
  Zhou, and Guoying Zhao.
\newblock Searching central difference convolutional networks for face
  anti-spoofing.
\newblock In {\em CVPR}, pages 5295--5305, 2020.

\bibitem{PLZero-Shot}
Xingxing Zhang, Shupeng Gui, Zhenfeng Zhu, Yao Zhao, and Ji~Liu.
\newblock Hierarchical prototype learning for zero-shot recognition.
\newblock {\em IEEE Transactions on Multimedia}, 2019.

\bibitem{CelebA-Spoof}
Yuanhan Zhang, Zhenfei Yin, Yidong Li, Guojun Yin, Junjie Yan, Jing Shao, and
  Ziwei Liu.
\newblock Celeba-spoof: Large-scale face anti-spoofing dataset with rich
  annotations.
\newblock In {\em ECCV}, 2020.

\bibitem{CASIA-MFSD}
Zhiwei Zhang, Junjie Yan, Sifei Liu, Zhen Lei, Dong Yi, and Stan~Z Li.
\newblock A face antispoofing database with diverse attacks.
\newblock In {\em ICB}, pages 26--31. IEEE, 2012.

\end{thebibliography}
}

\end{document}


\section{Notations}
\noindent In this part, we list all notations we used in this paper.
\setlength{\tabcolsep}{5pt}
\begin{table}[h!]
\centering
\ra{1.2}
\label{table:notations}
\resizebox{0.45\textwidth}{!}{%
\begin{tabular}{@{}ll@{}}
\toprule
Notation & Meaning \\ \midrule
$\mathcal{P}_j^r$ & The $r$-th prototype function of class $j$ \\
$f_i$ & The embedding of the $i$-th input data \\
$\theta_j$ & The angle between $x_i$ and the prototype distribution of class $j$ \\
$\mathcal{F}_j$ & A set of sample embedding of class $j$ \\
$t_j$ & Density threshold for class $j$ in the APS algorithm \\
$x_i$ & Given the $i$-th input data \\
$y_i$ & The label of the $i$-th input data \\
${\{ \mathcal{S}_k^s\}}_{k=1}^n$ & A set of spoof type labels of the input data, which is indexed by $k$  \\
${\{ \mathcal{S}_k^i\}}_{k=1}^n$ & A set of illumination labels of the input data, which is indexed by $k$ \\
$\mathcal{E}_j^\epsilon$ & The sample density of $\mathcal{P}_j^\epsilon$ in the APS algorithm \\
$\mathcal{D}_j^\epsilon$ & A set of samples within the region of $\mathcal{P}_j^\epsilon$ in the APS algorithm \\
$K$ & The number of prototype functions for the Live and Spoof class \\
$K_L$ & The number of prototype functions for the Live class \\
$K_S$ & The number of prototype functions for the Spoof class  \\
$\tau$ & The scaling factor of sample embedding in class prediction \\
$m$ & The angular margin penalty in $\mathcal{L}_{PD}$ \\
$s$ & The scaling factor of class prediction in $\mathcal{L}_{PD}$\\
$\delta_1$ & The margin penalty in $\mathcal{L}_{PC_{inter}}$ \\
$\delta_2$ & The margin penalty in $\mathcal{L}_{PC_{intra}}$ \\

\bottomrule
\end{tabular}%
}
\end{table}

\setlength{\tabcolsep}{5pt}
\begin{table}[t]
\ra{1.0}
\caption{Ablation study on training hyper-parameters.}
\vspace{3pt}
\label{table:training hyper-parameters}
\resizebox{0.45\textwidth}{!}{%
\begin{tabular}{@{}cccccccccc@{}}
\toprule 
&
\multicolumn{3}{c}{Hyper-params } &
  \multicolumn{3}{c}{TPR (\%)$\uparrow$ } &
  \multirow{2}{*}{APCER (\%)$\downarrow$ } &
  \multirow{2}{*}{BPCER (\%)$\downarrow$ } &
  \multirow{2}{*}{ACER (\%)$\downarrow$ } \\ \cmidrule(lr){2-4} \cmidrule(lr){5-7}
& $s$  & $\tau$  &  $m$    & FPR = 1\% & FPR = 0.5\%   & FPR = 0.1\%   &        &       &           \\ \midrule
\multirow{3}{*}{(1)} & 32.0 & 0.1 & 0.7 &
98.8 & 
97.0 & 
87.2          & 
0.98  & 
1.15  & 
1.06          \\ 
& 64.0 & 0.1 & 0.7 &
\textbf{99.5} & 
\textbf{98.4} & 
\textbf{90.3}          & 
\textbf{0.57}  & 
\textbf{0.94}  & 
\textbf{0.75}          \\ 
& 128.0 & 0.1 & 0.7 &
99.4 & 
98.1 & 
88.1 & 
0.67  & 
0.94  & 
0.80          \\ \hline
\multirow{3}{*}{(2)} &  64.0 & 0.01 & 0.7 &
99.2 & 
98.1 & 
\textbf{91.2} & 
0.88  & 
\textbf{0.82}  & 
0.85          \\  
& 64.0 & 0.1 & 0.7 &
\textbf{99.5} & 
\textbf{98.4} & 
90.3          & 
\textbf{0.57}  & 
0.94  & 
\textbf{0.75}          \\ 
& 64.0 & 1.0 & 0.7 &
99.2 & 
98.0 & 
89.9 & 
0.82  & 
0.92  & 
0.87          \\ \hline
\multirow{3}{*}{(3)} &  64.0 & 0.1 & 0.3 &
98.9 & 
97.3 & 
90.5 & 
0.93  & 
1.10  & 
1.01          \\ 
& 64.0 & 0.1 & 0.5 &
99.4 & 
97.7 & 
\textbf{91.0} & 
0.82  & 
0.92  & 
0.87          \\ 
& 64.0 & 0.1 & 0.7 &
\textbf{99.5} & 
\textbf{98.4} & 
90.3          & 
\textbf{0.57}  & 
\textbf{0.94}  & 
\textbf{0.75}          \\
\bottomrule 
\end{tabular}%
}
\vspace{0pt}
\end{table} 

\setlength{\tabcolsep}{5pt}
\begin{table}[htb!]
\scriptsize
\centering
\ra{1.3}
\caption{Results of cross-domain test on SiW. \textbf{Bolds} are the best results.}
\label{table:result of Cross-Domain test on siw}
\vspace{3pt}
\begin{tabular}{@{}ccccc@{}}
\toprule
Prot.              & Methods        & APCER(\%)$\downarrow$ & BPCER(\%)$\downarrow$ & ACER(\%)$\downarrow$         \\ \midrule
\multirow{6}{*}{1} & Auxiliary~\cite{lyjauxuliary}      & 3.58       & 3.58      & 3.58           \\
                  & FAS-SGTD~\cite{wang2020FAS-SGTD}         & 0.64       & 0.17    &   0.40 		\\
                  & BASN~\cite{KimBASN}           & -         & -       & 0.37              \\
                  & BCN~\cite{yztHMP}           & 0.55        & 0.17       & 0.36              \\
                  & CDCN~\cite{yztCDC}           & 0.07       & 0.17       & \textbf{0.12}    \\
                  & \textbf{LDA} &   0.14     &    0.33    & 0.24     \\
 \midrule
\multirow{6}{*}{2} & Auxiliary~\cite{lyjauxuliary}     & 0.57 $\pm $ 0.60  &  0.57 $\pm $ 0.60 & 0.57 $\pm $ 0.60                           \\

& BASN~\cite{KimBASN}          & -       & -      & 0.12 $\pm $ 0.03 \\

& BCN~\cite{yztHMP}   & 0.08 $\pm $ 0.17 &	1.15 $\pm$ 0.00	& 0.11 $\pm $ 0.08              \\

& CDCN~\cite{yztCDC}	& 0.00 $\pm $ 0.00 & 0.13 $\pm $ 0.09 &	0.06 $\pm $ 0.04     \\ 

 & FAS-SGTD~\cite{wang2020FAS-SGTD} &	0.00 $\pm $ 0.00	& 0.04 $\pm $ 0.08 & 	\textbf{0.02 $\pm $ 0.04} \\
 
 & \textbf{LDA} &   0.10 $\pm $ 0.03    &   0.14 $\pm $ 0.04    & 0.12 $\pm $ 0.02     \\
 

 \midrule

\multirow{6}{*}{3} & Auxiliary~\cite{lyjauxuliary}      & 8.31 $\pm $ 3.81   &  8.31 $\pm $ 3.81    & 8.31 $\pm $ 3.81          \\
& BASN~\cite{KimBASN}       & -   & -   & 6.45 $\pm $ 1.80          \\
 & FAS-SGTD~\cite{wang2020FAS-SGTD}	&  2.63 $\pm $ 3.72 &	2.92 $\pm $ 3.42 &	2.78 $\pm $ 3.57 \\ 
& BCN~\cite{yztHMP}           & 2.55 $\pm $ 0.89   &  2.34 $\pm $ 0.47   & 2.45 $\pm $ 0.68          \\
& CDCN~\cite{yztCDC}    &       1.67 $\pm $ 0.11	 & 1.76 $\pm $ 0.12& \textbf{1.71 $\pm $ 0.10}        \\ 
& \textbf{LDA} &  2.96 $\pm $ 1.42  &  2.23 $\pm $ 1.05     &   2.59 $\pm $ 1.24  \\
\bottomrule
\end{tabular}%
\end{table}

\section{Implementations Details}
\noindent For the concern of difficulty of implementation, we will release the code as soon as possible with the acceptance of the paper.

\noindent \textbf{Detailed training procedures.}
During the training stage, the prototypes are initialized with the random uniform distribution. They are jointly trained with the backbone and updated with the gradient from LDA/LDA$_{\mathcal{S}}$ loss. During the inference stage, which is shown in Fig.~2 in the paper, we utilize APS algorithm to remove redundant prototype functions adaptively by considering all the training samples. Then, the rest of the prototype functions are employed for final prediction. 

\noindent \textbf{Hyper-parameters of  the  loss  functions.}
As for experiments on Oulu-NPU and SiW, we set $\tau$ as 10.0, $s$ as 64.0, $\lambda_1$ as 0.1, $\lambda_2$ as 0.001, $\delta_1$ as 0.5, $\delta_2$ as 0.0, and $m$ as 0.7. As for semantic auxiliary for LDA on Oulu-NPU, the balanced parameter for spoof type is set to 0.1 and the balanced parameter for auxiliary session task is set to 0.01. As for the experiments on CelebA Spoof, we set $\tau$ as 10.0, $s$ as 64.0, $\lambda_1$ as 0.1, $\lambda_2$ as 0.001, $\delta_2$ as 0.0, and $m$ as 0.7. Besides, $\delta_1$ is selected from a list, which consists of 0.3, 0.5, and 0.7. The balanced parameters for the auxiliary task in terms of the live face attributes, spoof type, and illumination conditions are set to 1.0, 0.1, and 0.001 by following AENet$_{\mathcal{C}, \mathcal{S}}$~\cite{CelebA-Spoof}. 
Moreover, in order to show the effect of the training hyper-parameters, we conduct an ablation study on the intra-dataset of CelebA-Spoof.
Table~\ref{table:training hyper-parameters} shows that the performance of LDA$_{\mathcal{S}}$ fluctuates over different training hyper-parameters (1 for scale $s$, 2 for temperature $\tau$, and 3 for margin $m$).

\section{Experiments}


\noindent \textbf{Cross-Domain Test on SiW.}
The proposed method Latent Distribution Adjusting (LDA) achieves comparable results on SiW as shown in Tabel~\ref{table:result of Cross-Domain test on siw}. 

\noindent \textbf{Performance on Heavier Model.}
To demonstrate the proposed method's generalization ability on different backbones, we evaluate the performance based on a heavier backbone, \ie~Xception~\cite{xception}. We conduct corresponding intra-datatset testing and cross-domain testing on CelebA-Spoof. Due to the benchmarks in CelebA-Spoof leverage semantic information, we follow this setting by utilizing LDA$_{\mathcal{S}}$. As for intra dataset testing, compared to AENet$_{\mathcal{C}, \mathcal{S}, \mathcal{G}}$, LDA$_{\mathcal{S}}$ improves 60.4\% for ACER and achieves comparable results for TPR@FPR, as shown in Table~\ref{table:the results of intra-dataset test on CelebA-Spoof}. The above results indicate the proposed method's brilliant overall capacity based on Xception on the large scale dataset. 
As for cross domain testing, compared to AENet$_{\mathcal{C}, \mathcal{S}, \mathcal{G}}$, LDA$_{\mathcal{S}}$ improves 28.1\% for ACER on protocol 1 and improves 71.0\% for ACER on protocol 2,
as shown in Table~\ref{table:the results of cross-domain test on CelebA-Spoof}.
Besides, the same significant improvement is achieved in terms of TPR@FPR on protocol 1. The above results indicate the proposed method's great generalization ability based on Xception on the larger scale dataset.

\setlength{\tabcolsep}{5pt}
\begin{table}[tb!]
\centering
\ra{1.0}
\caption{Results of the intra-dataset test on CelebA-Spoof. \textbf{Bolds} are the best results.}
\label{table:the results of intra-dataset test on CelebA-Spoof}
\vspace{3pt}
\resizebox{0.45\textwidth}{!}{%
\begin{tabular}{@{}ccccccc@{}}
\toprule 
\multirow{2}{*}{Methods} &
  \multicolumn{3}{c}{TPR (\%)$\uparrow$ } &

  \multirow{2}{*}{APCER (\%)$\downarrow$ } &
  \multirow{2}{*}{BPCER (\%)$\downarrow$ } &
  \multirow{2}{*}{ACER (\%)$\downarrow$ } \\ \cmidrule{2-4}
   & FPR = 1\% & FPR = 0.5\%   & FPR = 0.1\%   &        &       &           \\ \midrule
AENet$_{\mathcal{C},\mathcal{S}, \mathcal{G}}$~\cite{CelebA-Spoof}       & 
\textbf{99.2} & 
98.4          & 
\textbf{94.2}          & 
3.72          & 
0.82         & 
2.27          \\
\textbf{LDA$_{\mathcal{S}}$} & 
98.8 & 
\textbf{98.7} & 
94.1         & 
\textbf{1.32}  & 
\textbf{0.51}  & 
\textbf{0.91}         \\                \bottomrule 
\end{tabular}%
}
\end{table}

\setlength{\tabcolsep}{5pt}
\begin{table}[tb!]
\centering
\ra{1.0}
\caption{Results of the cross-domain test on CelebA-Spoof. \textbf{Bolds} are the best results.}
\label{table:the results of cross-domain test on CelebA-Spoof}
\vspace{3pt}
\resizebox{0.45\textwidth}{!}{%
\begin{tabular}{cccccccccc}
\toprule
\multirow{2}{*}{Prot.} &
  \multirow{2}{*}{Methods} &
  \multicolumn{3}{c}{TPR (\%) $\uparrow$} &
  \multirow{2}{*}{APCER (\%)$\downarrow$} &
  \multirow{2}{*}{BPCER (\%)$\downarrow$} &
  \multirow{2}{*}{ACER (\%)$\downarrow$} \\ \cmidrule{3-5}
 &
   &
  FPR = 1\% &
  FPR = 0.5\% &
  FPR = 0.1\% &
   &
   &
   \\ \midrule
   
\multirow{2}{*}{1} 
&
AENet$_{\mathcal{C}, \mathcal{S}, \mathcal{G}}$~\cite{CelebA-Spoof} &
  96.9 &
  93.0 &
  83.5 &
  3.00 &
  1.48 &
  2.24  \\ 
& \textbf{LDA$_{\mathcal{S}}$}&
  \textbf{97.7} &
  \textbf{96.9} &
  \textbf{88.6} &
  \textbf{2.48} &
  \textbf{0.74} &
  \textbf{1.61}  \\  \midrule

\multirow{2}{*}{2} 
&
AENet$_{\mathcal{C}, \mathcal{S}, \mathcal{G}}$~\cite{CelebA-Spoof} &
  
  \# &
  \# &
  \# &
  4.77$\pm $4.12 &
  1.23$\pm $1.06 &
  3.00$\pm $2.90  \\
& \textbf{LDA$_{\mathcal{S}}$} &
  \# &
  \# &
  \# &
  \textbf{1.22$\pm$0.81} &
  \textbf{0.50$\pm $0.22} &
  \textbf{0.87$\pm$0.43}  \\
  \bottomrule
\end{tabular}%
}
\end{table}

\noindent \textbf{Ablation study on the number of initialized prototypes.} 
As shown in Table~\ref{table:performance of APS algorithm in LDA}, we conduct the ablation study on the intra-dataset benchmark of CelebA-Spoof to show the impacts of the numbers of initialized prototypes on the model performance. $K1$$\rightarrow$$K2$ represents the number of prototype variation which is provided by APS algorithm. 
We have two observations: 1) Different initialized numbers of prototypes cannot influence the final prototypes with APS algorithm, which shows the algorithm does not sensitive to the initialization, and 2) Larger number of initialized prototypes shows better performance. 
Besides, due to the small scale of academic datasets in terms of quantity and diversity, the model only needs few prototypes. We also conduct experiments on the internal larger scale (tens of millions) dataset and find more prototypes are required.

\setlength{\tabcolsep}{5pt}
\begin{table}[tb!]
\footnotesize
\centering
\ra{1.0}
\caption{Results of APS with different initialized prototypes.} 
\vspace{3pt}
\label{table:performance of APS algorithm in LDA}
\begin{tabular}{@{}ccccc@{}}
\hline
 & $K1 \rightarrow K2$ & ACER(\%) $\downarrow$ \\ \midrule
 & 2 $\rightarrow$ 2 &  1.13 $\rightarrow$ 1.13 \\
 & 4 $\rightarrow$ 3 &  0.98 $\rightarrow$ 0.99 \\
 & 100 $\rightarrow$ 3 &  0.89 $\rightarrow$ 0.88 \\
 & 200 $\rightarrow$ 3 &  0.88 $\rightarrow$ 0.89 \\
 \bottomrule [\heavyrulewidth]
\end{tabular}%
\vspace{-15pt}
\end{table}

